\pdfoutput=1

\documentclass[11pt,svgnames,table]{article}

\usepackage[]{arxiv}

\usepackage{times}
\usepackage{latexsym}

\usepackage[T1]{fontenc}
\usepackage[utf8]{inputenc}

\usepackage{csquotes}

\usepackage{microtype}

\usepackage{inconsolata}

\usepackage{listings}
\usepackage{cprotect}
\usepackage{multirow}
\usepackage{tabulary}
\usepackage{booktabs}

\usepackage{graphicx}
\usepackage{subfigure}

\usepackage[many]{tcolorbox}

\renewcommand{\mkbegdispquote}[2]{\fontfamily{cmr}\fontsize{10pt}{10pt}\selectfont}

\definecolor{lightblue}{HTML}{ff7f2a}
\definecolor{lighterblue}{HTML}{ffe6d5}
\newtcolorbox{mybox}[1]{colback=lighterblue,colframe=lightblue,title=#1}
\newtcolorbox{mybreakbox}[1]{colback=lighterblue,colframe=lightblue,breakable,title=#1}

\colorlet{key}{blue}
\colorlet{punct}{red!60!black}
\definecolor{background}{HTML}{EEEEEE}
\definecolor{delim}{RGB}{20,105,176}
\colorlet{numb}{magenta!60!black}

\lstdefinelanguage{json}{
    basicstyle=\small\ttfamily,
    numbers=left,
    numberstyle=\scriptsize,
    stepnumber=1,
    numbersep=8pt,
    showstringspaces=false,
    breaklines=true,
    frame=single,
    backgroundcolor=\color{background},
    literate=
      {:}{{{\color{punct}{:}}}}{1}
      {,}{{{\color{punct}{,}}}}{1}
      {\{}{{{\color{delim}{\{}}}}{1}
      {\}}{{{\color{delim}{\}}}}}{1}
      {[}{{{\color{delim}{[}}}}{1}
      {]}{{{\color{delim}{]}}}}{1},
}

\title{ImPaKT: A Dataset for Open-Schema Knowledge Base Construction}

  \author{Luke Vilnis   \quad Zach Fisher \quad Bhargav Kanagal \quad Patrick Murray \quad Sumit Sanghai\\
        Google Research \\
        \texttt{\{lvilnis,zachfisher,bhargav,pcmurray,sumitsanghai\}@google.com}}

\begin{document}
\maketitle
\begin{abstract}
Large language models have ushered in a golden age of semantic parsing. The seq2seq paradigm allows for open-schema and abstractive attribute and relation extraction given only small amounts of finetuning data. Language model pretraining has simultaneously enabled great strides in natural language inference, reasoning about entailment and implication in free text. These advances motivate us to construct ImPaKT, a dataset for open-schema information extraction, consisting of around 2500 text snippets from the C4 corpus, in the shopping domain (product buying guides), professionally annotated with extracted attributes, types, attribute summaries (attribute schema discovery from idiosyncratic text), many-to-one relations between compound and atomic attributes, and implication relations. We release this data in hope that it will be useful in fine tuning semantic parsers for information extraction and knowledge base construction across a variety of domains. We evaluate the power of this approach by fine-tuning the open source UL2 language model on a subset of the dataset, extracting a set of implication relations from a corpus of product buying guides, and conducting human evaluations of the resulting predictions.
\end{abstract}

\section{Introduction}

\begin{figure}[t!]
        \includegraphics[width=0.5\textwidth]{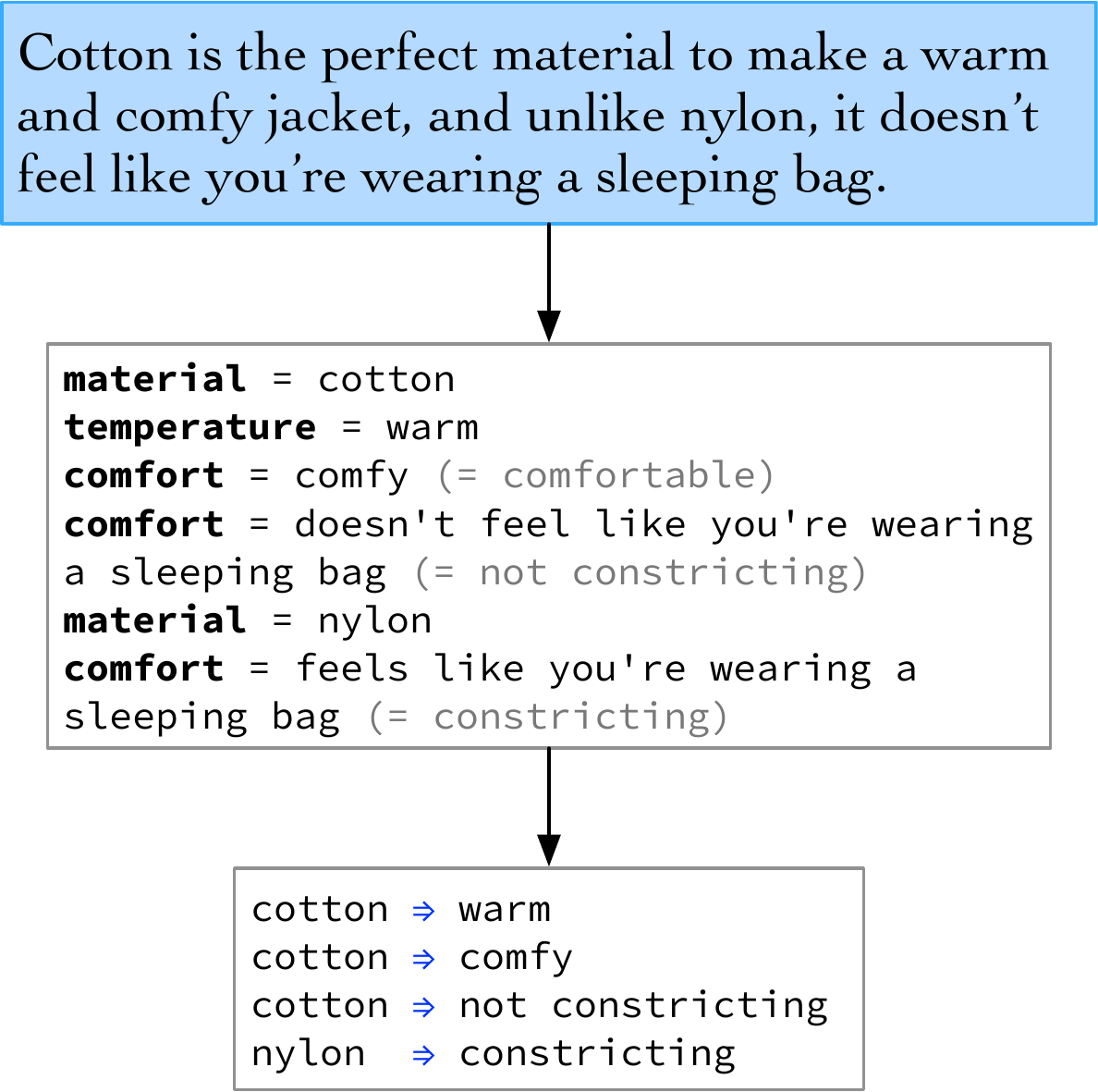}
    \caption{Parsing a semi-structured implication relation from web text. We annotate attributes, and their summarized forms (in gray), attribute types (in bold), and implication relationships.}
 \label{fig:wine-implications}
\end{figure}

Knowledge bases (KBs) have long been important components of machine learning systems for natural language understanding and information synthesis, from commonsense knowledge bases like Cyc \cite{lenat1995cyc} and ConceptNet \cite{speer2017conceptnet}, to general KBs like FreeBase \cite{freebase:datadumps},  DBpedia \cite{10.5555/1785162.1785216}, Wikidata \cite{10.1145/2629489}, YAGO \cite{10.1145/1242572.1242667}, or domain-specific KBs like the medical UMLS KB \cite{bodenreider2004unified}.

Knowledge of structured relations, entities, and attributes is important for providing reliable grounding for downstream tasks like question answering \cite{https://doi.org/10.48550/arxiv.2209.04994}, summarization \cite{https://doi.org/10.48550/arxiv.2104.07606} and faceted search~\cite{tunkelang2009faceted} and recommendations~\cite{sun2018conversational}. In commercial settings, such as the shopping domain \cite{yang2022mave}, it is useful for understanding user preferences and use cases for different products, and how features interact.

As machine learning is applied more deeply and to more domains, manually curated knowledge bases become less adequate for the task, and automated knowledge base construction (AKBC) becomes a more important problem than ever. Additionally, the fixed schemas of attribute types and relations in many traditional knowledge bases prove brittle and inflexible in their application to new domains. Approaches to open-schema KBC such as OpenIE \citep{openie}, NELL \citep{nell}, and Universal Schema \citep{riedel-etal-2013-relation} are more important than ever in domains such as common sense reasoning, medical, shopping, materials science, and others with an ever-evolving set of attributes, types, and relationships. Finding the mentions, attributes, relations for these open schemas can be difficult though and relies on human labeling, syntactic rules, and heuristics like Hearst patterns \cite{hearst-1992-automatic}. For example, OpenIE and Universal Schema both rely on pre-trained dependency parsers to extract syntactic substructures as candidates for relations and types.

Recently, neural networks, and specifically Large Language Models (LLMs) such as GPT-3 \cite{brown2020language} and T5 \cite{raffel2020exploring,xue2020mt5} have ushered in a golden age of semantic parsing. These models are capable of extracting semistructured and structured data from freeform text with remarkably little supervision, and the seq2seq paradigm \cite{sutskever2014sequence} enables us to construct structured annotations through simple text-based templating \cite{lu2022unified}.

As open information extraction enters the LLM age, we are inspired to release the ImPaKT (Implication Parsing and Knowledge Extraction) dataset, a dataset for training semantic parsers for information extraction and open-schema semistructured AKBC. Our dataset consists of thousands of annotations of fine-grained attributes, types, and implication relationships on C4 \cite{raffel2020exploring} sentences from the shopping domain, but we believe it will be both useful training data and a useful annotation template for constructing KBs in many domains. In addition to being a commercially valuable domain to understand in its own right, the shopping domain possess qualities such as fine grained, ever evolving attribute type schemas, and common-sense style relations, that make it ideal testbed for many annotation patterns of interest.

While many existing datasets for attribute extraction exist, ours is uniquely tailored for the strengths of LLMs. Attribute extraction datasets like MAVE \cite{yang2022mave} and MAE \cite{DBLP:journals/corr/abs-1711-11118} have a large number of typed shopping-related attributes, but they correspond strictly to text spans, and lack the mapping of raw or loosely paraphrased text to the human-created summarized forms that we provide. This allows models trained on our data to have higher recall and even to do schema discovery in new domains. Further, they lack the many-to-one associations between compound and atomic attributes that allow us to parse complex relations. Commonsense datasets like ConceptNet may contain some relations reminiscent of our implications, but they lack the rest of our context and provenance, and are noisy.

We choose to annotate implication relationships because they possess many qualities of commonsense reasoning, a long standing goal of natural language understanding, as well as being valuable to understand in the shopping domain for applications such as understanding consumer use cases e.g. understanding that a cell phone with \emph{large buttons} would also be a cell phone that is \emph{good for seniors} to use. Parsing implication relationships shares many similarities with the natural language inference  (NLI) \citep{maccartney2009natural,bowman2015large} and recognizing textual entailment (RTE) \citep{baroni2012entailment,dagan2013recognizing}, wherein entailment (implicative) relations between sentences or concepts are classified, a problem of longstanding interest to the NLU community. Implication relationships themselves are also useful for performing multi-hop reasoning and discovering causal chains between attributes, an active area of research in KG completion \cite{das2016chains,das2017go}.

We demonstrate the utility of our dataset in training information extraction models by finetuning a UL2 language model on a mixture of seq2seq tasks including attribute discovery, annotation, and implication extraction, and conduct a human study of the resulting knowledge base. We release our dataset publicly in hopes that researchers will use it to train their own parsers, construct their own fine grained knowledge bases, and advance the study of automated semistructured knowledge base construction.

\section{Related Work}

Machine-learned open-schema information extraction has a long history, pioneered by works such as OpenIE \citep{openie}, NELL \citep{nell}, and Universal Schema \citep{riedel-etal-2013-relation}. 

Our work in open-schema and fine grained semistructured semantic parsing, with an emphasis on common nouns, shares a lot of similarities with the literature on commonsense and commonsense knowledge base construction, a fact that has been noted by others studying information extraction in the shopping domain \citep{https://doi.org/10.48550/arxiv.2211.08316}. Early systems like Cyc \citep{lenat1995cyc} rely on a manually curated set of relations between common nouns, while later work such as NELL \citep{nell} and ConceptNet has included more automatic extractions \citep{speer2017conceptnet}. More recently, neural networks have been applied to commonsense KB completion \citep{li2016commonsense} and now with the rise of large language models, there is considerable research interest in the kind of commonsense knowledge implicitly learned in the weights of these models \citep{li2021systematic}. To this end, recent work has attempted to explicitly construct a knowledge base using the latent knowledge in the language model \citep{hao2022bertnet}. 

In contrast to this work, which has mostly focused on KB completion and implicit knowledge, our work brings large language models to bear on the problem of explicit information extraction. In many domains, we wish to parse textually well-supported statements from a corpus, and leave inferences for a separate part of the KB pipeline. Recent work in the shopping domain has extracted large sets of attributes and types from Amazon data \citep{yang2022mave}.  We extend this approach with a more fine-grained annotation scheme including attribute summaries, compound and atomic attributes, and implication relationships, which we believe is suited for even more domains and applications within the shopping domain, such as recommendation and summarization \citep{https://doi.org/10.48550/arxiv.2104.07606,https://doi.org/10.48550/arxiv.2212.01956}.

The implication relationships we annotate share similarities with the literature on natural language inference (NLI) \citep{maccartney2009natural,bowman2015large} and recognizing textual entailment (RTE) \citep{baroni2012entailment,dagan2013recognizing}. In contrast to this work, which aims to find when sentences or concepts are logically entailed from one another (performing a sort of inference over implicit knowledge), we focus on parsing textually grounded statements where the text itself asserts an implication relationship.

Neural networks have long been employed for structured prediction and information extraction tasks such as semantic and syntactic parsing, often using custom architectures such as graph neural networks \cite{DBLP:journals/corr/abs-1806-01261} or pointer networks \cite{https://doi.org/10.48550/arxiv.1506.03134}. Like much other recent work, our proposed semantic parsing model follows the seq2seq framework \citep{sutskever2014sequence} in encoding input and output structures as simple textual sequences encoded with a sentencepiece model, like recent work in sequence tagging with transformers \citep{https://doi.org/10.48550/arxiv.2203.08378}, event extraction \cite{hsu2022degree}, and autoregressive entity retrieval \citep{DBLP:journals/corr/abs-2010-00904}. Our use of a single LLM to generate multiple structured annotations is in the vein of the Universal Information Extraction model of \citet{lu2022unified}.

\section{Structure of the ImPaKT Dataset}

\begin{table*}[t]
    \centering
    \begin{tabular}{*6c}
        \toprule
        & \multicolumn{5}{c}{Annotation Aspects} \\
        \cmidrule(lr){2-6} 
        & Snippets & Coarse Attribute & Implication & Atomic Attribute & Atomic Implication     \\
        \midrule
        Count & 2489 & 5655 & 3719 & 6117 & 3587 \\
        \bottomrule
    \end{tabular}
    
    \caption{Count of annotation aspects in dataset. Each aspect of annotation has potentially multiple annotations associated with it. Snippets are labeled with GPC categories and classification decisions (1628 positively containing an implication, 859 negatives either for reason of being malformed text, in the wrong category, or containing no implications). Coarse attributes are further labeled with summaries. Atomic attributes are also labeled with summaries and types. }
 \label{tab:annotation-counts}
\end{table*}

\begin{figure*}
\centering
\begin{minipage}{\textwidth}

\begin{lstlisting}[language=json,,numbers=none]
  "snippet": 
    "Uses durable 7075 aluminum stakes that won't succumb to conditions",
  "provenance": {
    "url": "https://www.gearhungry.com/best-solo-tents-for-camping/",
    "timestamp": "2019-04-19T21:11:49Z",
    "span_start": 1531,
    "span_end": 1597
  },
  "category": 
    "Sporting Goods > Outdoor Recreation > Camping & Hiking > Tents",
  "classification": "Yes",
  "attributes": [
    {
      "attribute": "durable 7075 aluminum stakes",
      "summary": "durable 7075 aluminum stakes"
    },
    {
      "attribute": "won't succumb to conditions",
      "summary": "impervious to conditions"
    }
  ],
  "atomic_attributes": {
    "durable 7075 aluminum stakes": [
      {
        "attribute": "durable",
        "summary": "durable",
        "attribute_type": "stake durability"
      },
      {
        "attribute": "aluminum",
        "summary": "aluminum",
        "attribute_type": "stake material"
      },
      {
        "attribute": "7075 aluminum",
        "summary": "7075 aluminum",
        "attribute_type": "stake aluminum alloy"
      }
    ],
    "won't succumb to conditions": [
      {
        "attribute": "won't succumb to conditions",
        "summary": "durable",
        "attribute_type": "stake durability"
      }
    ]
  },
  "implications": [
    {
      "antecedent": "durable 7075 aluminum stakes",
      "consequent": "won't succumb to conditions"
    }
  ]
\end{lstlisting}
\end{minipage}
\caption{Example annotation. In this case, several atomic attributes correspond to one compound antecedent in an implication.}
\label{fig:sample-annotation}
\end{figure*}

In creating this dataset, we had to reconcile the annotation of fine-grained structure with the intrinsic ambiguity of natural language and human semantic judgments, as well as making it possible for human annotators to provide reasonable labels. To that end, we split up some of the annotations into several levels of granularity. The attributes come in two varieties, general (possibly \emph{compound}) attributes and \emph{atomic} attributes. All attributes are represented by a string name and a possibly shorter and more normalized name called a \emph{summary}. The atomic attributes further come with a type annotation.

Since our dataset is constructed in the consumer product domain, our examples will also come from that domain. These annotation principles should carry over to many domains that can benefit from semantic parsing of semistructured knowledge.

\subsection{Attributes}

In this work we refer to an \emph{attribute}, or \emph{coarse} attribute when context requires disambiguation, as a string describing any aspect of a product in a particular category, relevant to the use of the product or the product category, and not necessarily based on any normalized schema. For example, \emph{good for large sandwiches} could be a perfectly good attribute describing a toaster oven or panini press. Clearly, the full set of attributes for any given product is potentially extremely large and sometimes dependent on the opinions of the users of that product, so in this work we do not attempt to construct a knowledge base of every possible attribute. Instead we annotate specific textual \emph{snippets} with particular choices of attributes, without duplicating concepts, and when in doubt how to phrase the particular attribute string we try to make it as close as possible to the phrasing in the snippet. 

Often times, the surface text of attributes is presented in a way that is not suitable for a more structured annotation. For this reason, we additionally annotate each attribute with a \emph{summary}. For example, for a snippet like \emph{a quality PC chair must offer proper neck and lumbar support, allowing the user to adopt an ergonomic posture}, we might lightly rephrase one attribute as \emph{allows user to adopt an ergonomic posture}, but we can further summarize this in context as simply \emph{ergonomic}.

\subsection{Implications}

In the context of our dataset, we define an \emph{implication} between two attributes $A$ and $B$ in the same category as a judgment that the existence of attribute $A$ describing a product also leads us to believe that attribute $B$ describes the product, \emph{grounded in a specific textual statement or statements}. That is, the text itself states the implication exists, we do not infer the implication. For example, the snippet \emph{the chair has a memory foam cushion to promote good posture} expresses an implication.

Either the left or right hand side of an implication might have several different components in conjunction, disjunction, or even some other fuzzy qualifiers. This is why we do not require our notion of attributes to correspond to one single crisp concept, or else we would be unable to annotate many implications of interest, such as \emph{these white Reebok tennis shoes are comfortable on a hot day}, which has individual aspects of color, brand, and type all bundled up into one side of the implication in an ambiguous manner.

\subsection{Atomic Attributes and Attribute Types}

While attributes are simply short descriptors of a product, in order to provide more finegrained semantic annotations, we consider the idea of an \emph{atomic attribute}. 

An atomic attribute is an attribute that further expresses only one concept of one type. Our coarse attributes may correspond to multiple atomic attributes, a mapping for which is provided in our dataset. Like coarse attributes, each atomic attribute comes with a summary, however unlike for coarse attributes, the summary might only convey the full information of the attribute in combination with the attributes \emph{type}, which is also provided. 

For example, in the golf-related snippet \emph{wedges tend to have the largest loft angles in a club set}, the attributes \emph{wedges} and \emph{largest loft angles} could be normalized into one choice of atomic attributes as \emph{golf club type = wedge} and \emph{loft angle size = largest} (where we write \emph{type = summary}). This gives us an annotation methodology to provide powerful semistructured open schemas that stand a good chance of mapping multiple source strings to the same attributes.

\subsection{Atomic Implications}

An atomic implication is an implication where both sides, antecedent and consequent, are atomic attributes. While the experiments in this paper are conducted only with respect to the atomic implications, we are excited to use this dataset to learn structures for more complex implications between compound attributes in future work (such as those found in Figure \ref{fig:sample-annotation}), and to see what the community does with this data as well. 

Statistics for each type of annotation in the dataset are found in Table \ref{tab:annotation-counts}, and a full annotated example is found in Figure \ref{fig:sample-annotation}.

\section{Data Collection}

\begin{figure*}
\centering
        \includegraphics[width=\textwidth]{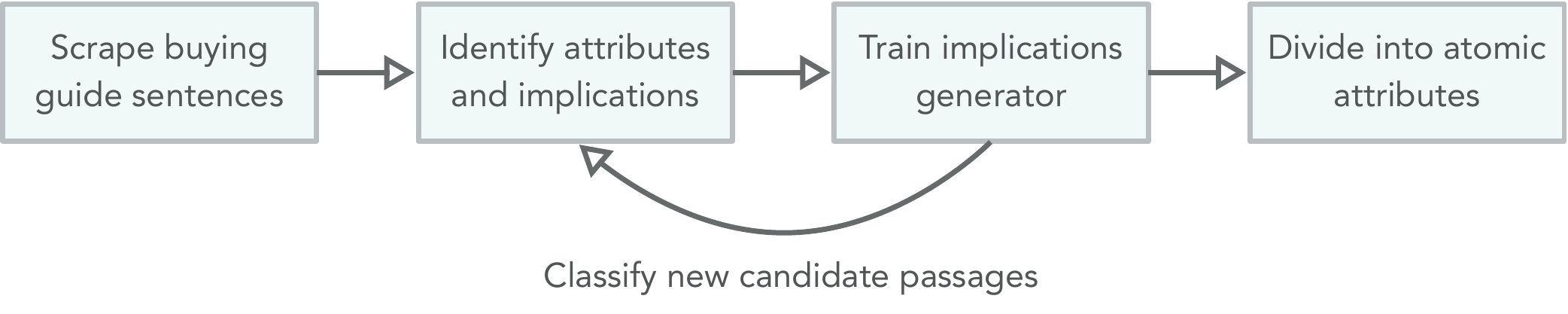}
\cprotect\caption{Annotation workflow for raters.}
\label{fig:annotation-pipeline}
\end{figure*}

The data were collected over a period of about six months from a team of 5-10 skilled professional annotators working on a variety of projects. Due to the inherent difficulty and ambiguity in providing these kind of semantic annotations, as well as finding candidate sentences to annotate, we conducted multiple rounds of annotation and correction cycles. What we describe here is an abridged version which captures the essential process, an overview of which is given in Figure \ref{fig:annotation-pipeline}.

\subsection{Candidate Generation}

Finding candidate sentences that contain implications between shopping attributes, with limited annotation resources, suffers from a cold-start problem. We chose to focus our efforts on \emph{buying guides}, web pages that compare and contrast products in various categories and offering recommendations on the best ones to purchase. We use a nonpublic corpus of candidate buying guide URLs, each automatically annotated with a product category from the publicly available Google Product Category taxonomy, and then further filter down this set of URLs with a regular expression \cprotect\footnote{We ue the following regex: \begin{lstlisting}[language=json,numbers=none]
(-|/)(guide(s)?|review(s)?
|buy(ing|s)?|vs|choos(e|ing)
|best(list(s)?)?|pros-cons|perfect
|top|most|spectacular|must-have
|cool(est)?|great(est)?
|exceptional|unique)(-|/|\.)
\end{lstlisting}}
in order to create a simple buying guide corpus. 

We split the pages into short sentences/snippets with a simple regex, took a sample of 500 sentences from these webpages, along with the category information, and asked raters to label them as containing an implication between attributes in that category (details given later). Using these 500 labels, we finetuned a pre-trained T5-like language model as a classifier to generate the rest of the candidates, sampling from the corpus a 70:30 ratio of sentences classified as containing an implication to sentences classified as not. During further rounds of data collection we continued to retrain the classifier in order to improve candidate generation.

\subsection{Coarse Annotation of Implications}

Given the set of category-annotated candidate snippets, we first asked annotators to annotate coarse, potentially compound, attributes and summaries of those attributes, as well as any implications between them. We asked the annotators to try their best to make the attributes atomic, except when making them compound was necessary to properly describe either side of an implication. We iterated on this task for a few rounds providing feedback, but keeping the annotation structure and documents quite similar, only adding examples of better and worse annotations. Relevant annotator instructions are provided in Appendix \ref{app:implication-parsing-instruction}.

\subsection{Fine-grained Annotation of Atomic Attributes}

We followed up by dividing those potentially compound attributes into atomic attributes, with summaries and attribute type annotations, and describing which of the previously annotated attributes each of the atomic attributes might be a sub-attribute of. Relevant instructions are provided in Appendix \ref{app:attribute-splitting-typing-instruction}.

\section{Experiments}

\begin{figure*}
\centering
        \includegraphics[width=\textwidth]{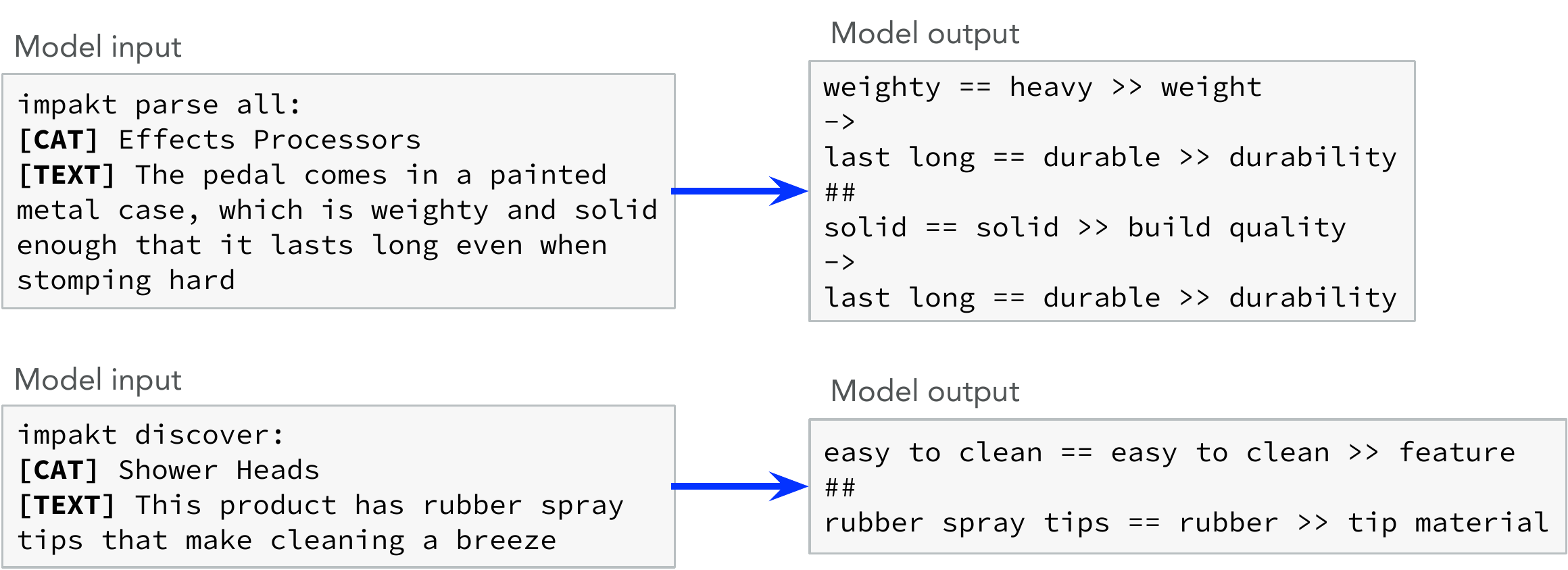}
\cprotect\caption{seq2seq encoding for attribute discovery and implication parsing. Discovered attributes are encoded as 
\verb|tag == summary >> facet|
and implications are encoded as \verb|attribute -> attribute|. Multiple annotations are separated with \verb|##|.}
\label{fig:seq2seq-encoding}
\end{figure*}

We conduct a holistic evaluation of our annotation dataset by training a parser and constructing a small knowledge base. Evaluating so called \emph{cold-start} knowledge bases with open schemas of entities and/or relations is well known to be quite difficult. To this end, we employ human raters to evaluate the precision of a small open KB of attribute implications created using a model trained on our data.

First, we restrict ourselves to set of annotated atomic attributes, and the set of \emph{atomic implications}, implications whose left and righthand sides correspond to only a single atomic attribute. We use these data to create training examples for several different tasks centered around predicting various aspects of attributes, summaries, types, and implications. The two relevant tasks are:

\begin{itemize}
    \item \textbf{Attribute discovery:} given a text snippet and a product category, produce the full set of atomic attributes contained in that snippet relevant to the product category, along with a summary and a type for each attribute.
    \item \textbf{Implication parsing:} given a text snippet and a product category, produce the set of implications between atomic attributes contained in that snippet relevant to the product category, along with a summary and a type for each.
\end{itemize}

For a base model, we choose the publicly available 20B parameter UL2 language model \citep{tay2022unifying} implemented in the T5X and SeqIO frameworks \citep{t5x}. 

We use a seq2seq encoding of our tasks using the mT5 vocabulary, which we demonstrate in Figure \ref{fig:seq2seq-encoding}. For multiple attributes or implications, we choose the order according to a fuzzy order of appearance in the input string given by a partial string overlap function. We use simple unconstrained greedy decoding to perform inference, and parse the outputs using a regular expression. Malformed spans of the output will simply fail to parse with the regex, and result in no annotation being produced for that span, but we find this does not happen in practice.

We finetune the model on our task mixture until the IID eval set converged, and then run inference with the attribute discovery and implication parsing tasks on a subsample of 1000000 sentences from the corpus. 
In order to construct our demonstration knowledge base, we aggregate these parses in two ways. For our parsed attributes, we discard the raw attributes and consider the attribute summary along with the type to be the canonical attribute: two attributes are the same if they have the same summary and type. For parsed implications, we consider two implications to be the same if both their left- and righthand sides match.

For our first experiment, we aggregate the parsed implications over the entire corpus and consider the count of times the implication appears as its confidence score, and take the top 100 most confident implications. We call these \emph{parsed implications}. 

\begin{figure*}[t!]
        \includegraphics[width=0.5\textwidth]{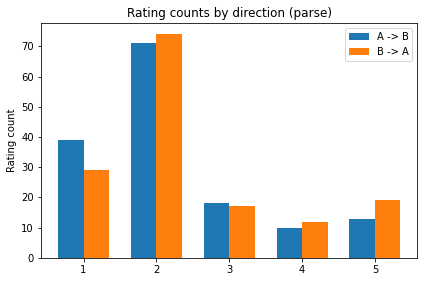}
    ~ 
        \includegraphics[width=0.5\textwidth]{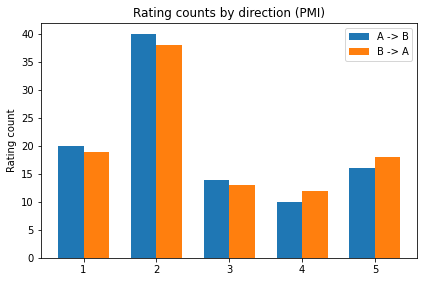}
    \cprotect\caption{Counts of rater evaluations for common implication candidates generated directly by the parser, or by PMI between parsed attributes. Candidates generated by the parser have stronger directionality and precision.
    (rating of \verb|A->B| is $1$, rating of \verb|B->A| is otherwise). }
 \label{fig:rater-counts}
\end{figure*}

\begin{table*}[t]
  \centering
\begin{tabular}{|>{\itshape}c|>{\itshape}c>{\itshape}c|>{\itshape}c>{\itshape}c|}
\hline
 \multirow{2}{*}{\normalfont{Category}} & \multicolumn{2}{c|}{Antecedent} & \multicolumn{2}{c|}{Consequent} \\ \cline{2-5}
 & \normalfont{Type} & \normalfont{Attribute Summary} & \normalfont{Type} & \normalfont{Attribute Summary} \\ \hline
Pillows & material & latex & naturality & natural \\
Bed Sheets & material & flannel & warmth & warm \\
Vitamins \& Supplements & nutrient & calcium & benefit & bone health \\
Shoes & upper material & mesh & durability & durable \\
Bicycles & material & aluminum & features & rust-resistant \\\hline
\end{tabular}
\cprotect\caption{Example of aggregated implications extracted from the corpus by the parser, judged by raters to have strong directionality (rating of \verb|A->B| is $1$, rating of \verb|B->A| is otherwise). }
 \label{tab:example-implications}
\end{table*}

For our second experiment, we aggregate counts of attributes and pairs of attributes, and calculate the pointwise mutual information between all pairs of attributes. This relationship does not discriminate directionally, but still captures many interesting implications. We call these \emph{PMI implications}.

To evaluate the precision of these knowledge bases without any notion of a ground truth schema, we employ an additional set of raters. The raters are asked to judge the quality of the parsed implications $A \to B$ as well as the reverse implication $B \to A$ using a Likert-style scale: 
\begin{mybox}{Instructions --- Implication precision}
Two properties, A and B, can describe a product in category C.
What is the most likely relationship out of the following?
\begin{enumerate}
    \item Given A, B must necessarily be true.
    \item Given A, B is likely to be true.
    \item A and B are usually not related.
    \item Given A, B is not likely to be true.
    \item I don’t have enough information or background.
\end{enumerate}
\end{mybox}

We provide the 100 examples from the parser and 100 from PMI to 9 raters, and duplicate each example 3 times to get multiple ratings for each. Because this is a difficult evaluation task, we selected only 4 of the 9 raters to include in our final evaluation, based on their understanding of some control questions for which we judged the ground truth as particularly clear. This left us with 100 annotations for the PMI implications, and 151 for the parsed implications.

We present the results of these experiments in Figure \ref{fig:rater-counts}. Blue bars indicate the judgments of raters of implications in one direction (in the case of parsing, this is direction indicated by the model), and orange bars indicate the judgments in the opposite direction of the model. In the case of PMI examples we can see that the implications learned are generally judged to be bidrectional, as would be expected. In the case of the parsed examples we see significantly more directionality for the implications labeled with option 1, that is, the judgment that if $A$ is true, $B$ must be true. For implications labeled with option 2, we see more of a lack of directionality, possibly because the judgment \emph{Given A, B is likely to be true} can be quite common in both directions for a variety of attributes of consumer products (e.g. attributes like \emph{lightweight}, \emph{portable}, \emph{breathable}, \emph{comfortable}, etc), and we are evaluating specifically the most commonly found implications in the corpus.

We provide examples in Table \ref{tab:example-implications} of some of the implications found by the parser which have the most directionality  according to the raters. While both the parser and PMI method appear able to learn many associations, valuable information for our open-schema knowledge base, the parser is higher precision and can further find reasonable directional implications between product attributes.

\section{Conclusion and Future Work}

In this work, we introduce a new dataset for training models to perform open-schema knowledge base construction in the shopping domain. We annotate thousands of free text types, attributes, and implication relationships. We demonstrate that high-precision implication relationships from text can be annotated using a UL2 language model finetuned on tasks from our corpus, both directly parsing the implications, and looking at PMI between pairs of parsed attributes. 

We release this data to the community in hope that researchers will use it to study not just shopping attribute and relation annotations, but also other domains that benefit from fine-grained free text, semistructured data, and open-schemas, such as commonsense knowledge. More broadly, we advocate for practitioners to take advantage of the tremendous power of large language models for semantic parsing, and to construct ambitious fine grained knowledge bases in a variety of domains.

\bibliography{paper}
\bibliographystyle{arxiv_natbib}

\newpage
\onecolumn
\appendix
\begin{center}
\Large{Appendix}
\end{center}

\section{Implication parsing annotator instructions}
\label{app:implication-parsing-instruction}

We provided raters with the following instructions for annotating implications:

\begin{displayquote}
We use the nomenclature of attributes and types. A attribute is a specific value that describes a product, for example “red” or “affordable”. A type is a key (i.e. a name) that describes a collection of related attributes. For example, the attribute “red” belongs to the type of “color,” and the attribute “affordable” belongs to the type of “price.”

In this task, we will ask you to identify the attributes in a snippet of text relevant to a specific provided product category that contain an \emph{Implies} relationship between them, as evidenced by the snippet. We wish to determine if a snippet is telling us that one attribute implies another, from information contained in that snippet. For example, “This beer is crisp and refreshing” does not tell us that there is any implication between the attributes “crisp” and “refreshing”, nor does it imply that all beers are crisp and refreshing, so we would mark this as “No implication”. However, “This computer has a fast GPU which is great for gaming” does tell us that the attribute “fast GPU” implies the attribute “great for gaming”. It should already be clear that there is a degree of fuzzy logical ambiguity here, since maybe not all fast GPUs are great for gaming, but in the context of the sentence this implication holds.

\textbf{Extracting attributes:} The first step is to identify the relevant implication-participating attributes mentioned in the provided snippet. In some cases there is ambiguity as to what the attribute is exactly, for example “the jacket is waterproof” could have the attribute “is waterproof” or just simply “waterproof”. Either choice is okay, but as a rule of thumb, the shortest text that captures the full meaning of the attribute is preferable. For example, “the refrigerator has cheap maintenance” might have the attribute “cheap maintenance,” as shortening this further to just “cheap” or “maintenance” loses the meaning of the attribute. Although a product category is provided, we would like to extract all attributes relevant to implications, regardless of category. For example, in the category of “Shoes”, given the snippet “These neutral color shoes look great with my red cocktail dress or other formal attire”, we could extract “neutral color” and “look great with red cocktail dress or other formal attire” as attributes related to the shoes that have an implied relationship. But we would also want to extract “cocktail dress” and “formal attire” because of the implied relationship between these two attributes even though the dress itself is not relevant to the category “Shoes”. 
The categories are only provided to give context in understanding the snippet, not to limit the scope of possible attributes and implications.

When generating attributes, they can be copied directly (from the provided snippet) when the copy makes sense on its own. However, in some cases, it may not make sense to directly copy; in this case, please add in the relevant text to make the attribute make sense on its own. For example, when generating numeric attributes, do your best to include units, even if they are not directly found in the text; in the phrase “The two possible screen sizes are 11 and 13 inches”, there should be a attribute for “11 inches” and one for “13 inches.” 

\textbf{Paraphrasing attributes:}
Sometimes you will need to paraphrase rather than copy the text directly. 

\begin{itemize}

\item  \textbf{Example 1:}  “I'm not going to write about the great tones and playability of this guitar, because its plays and sounds like any other \$200 guitar on the market.” In this case, you would not want the attribute to be “great tones'' since the user is essentially saying that the tones are average and so is the playability. So go ahead and paraphrase the attributes as ``Average tones'' and ``Average playability''  and mark an X in the Paraphrase column.

\item  \textbf{Example 2:} Be concise when paraphrasing, but aim to keep the specificity of the original text. Text passage:
\begin{displayquote}
 That said, there are currently two major disadvantages we've experienced with some of the Mirrorless cameras we've encountered so far. The first is the inability of many models to match the autofocus speed of DSLRs when conducting continuous or predictive AF 
\end{displayquote}
Rather than “slower autofocus”, the attribute should be more specific and therefore include more information: “slower autofocus speed than DSLRs when conducting continuous or predictive AF” is the preferred attribute.

\end{itemize}

Do not alter the text snippets in the text column. However, if you're extracting/paraphrasing a attribute with a misspelling, please spell the word(s) correctly in the attribute column. 

\textbf{Marking implications:} In this task, we are interested in finding implication relations between attributes. Given a snippet, we want to establish whether the text of the snippet itself (along with some basic “commonsense” about the product category and any other context that an average person could be expected to know) gives reason to believe one of these implications holds between the attributes.

For two attributes A and B from snippet S, the existence of the relation Implies(A, B) means that given the information in S, we can infer that the attribute A implies the attribute B, perhaps not in a strict logical sense, but in context. For example, the sentence “This computer has a fast GPU which is great for gaming” would give us the relevant (implication-participating) attributes “fast GPU” and “great for gaming”, along with the relation Implies(fast GPU, great for gaming). This might not be a strict logical implication (what if the fast GPU overheats and makes it hard to play games?), but in the context of the sentence, this implication holds.

\textbf{``Is-'':} A subset of implications come about strictly because of a straightforward is-a / sub-tag relationship between the attributes. For example, the snippet “Wool is very good for casual wear as it doesn't smell and merino wool is very soft and comfortable” could give us the relevant attributes “wool”, “good for casual wear”, “doesn’t smell”, “soft” and “comfortable”, along with the relations Implies(wool, good for casual wear), Implies(wool, doesn’t smell), Implies(merino wool, soft), Implies(merino wool, comfortable), and Implies(merino wool, wool). This latter implication, Implies(merino wool, wool), can be inferred from the sentence, and you should feel free to mark it, but if it becomes distracting to figure out too many of these implied taxonomic sort of relationships, do not worry too much about missing them. 

\textbf{Implied Negations:} Some snippets will imply a negation of one of the attributes, for example, in a snippet like “The priciest jeans are made with high-quality cotton, while bargain brands will often use a synthetic mix”, we can infer that “high-quality cotton” implies not “synthetic mix” in the context of clothing and jeans. We can denote this as Implies(high-quality cotton, NOT synthetic mix) with the “NOT” capitalized. The capitalization denotes that we added this “NOT” as an inference step, as opposed to seeing the “not” directly in the text, but if you are confused between whether the “NOT” should be capitalized, it is not very important and it is ok to make it lower case.

\textbf{Implicit ANDs vs implicit ORs:} in the example above, we saw that a group of attributes were implicitly related by an OR in the context of the implication. This is not always the case though, sometimes a whole group of things are required to be present at once in order for the implication relationship to hold. For example, given the snippet “For summer season, a muslin cotton swaddling blanket is always a perfect choice”, we can infer that a “swaddling blanket” is not always a perfect choice for summer, it must be a “muslin cotton swaddling blanket”. So we could write the implication Implies(muslin cotton swaddling blanket, perfect for summer season). It would also be acceptable to write Implies(muslin AND cotton AND swaddling blanket, perfect for summer season), but breaking out the “logical operator” like this is not nearly as important in the case of ANDs as it is in the case of ORs, since in the latter we are using OR as simply a way to concisely write down a whole group of relations and make annotation easier.

\textbf{Bad examples and “no implications”:} There are two ways for a given snippet to not have any useful implications to extract. In one case, the snippet might be a meaningful sentence but simply not give evidence of any implication relationship. In that case, we mark the example as “No implication”. Some snippets, on the other hand, are just “bad examples”, meaning that they are malformed in some way such as not being meaningful sentences. A common way for that to happen is with product titles — our data is extracted automatically and some of the extracted snippets are simply titles of products, like “2019 New Trend Fashion Summer Solid Color Shirt Men Stand Collar Cotton Linen Half Sleeve Shirt Slim Type Male Casual Linen Tops”. This should simply be marked as “Bad example”. In early iterations of this task, there might be a fair amount of bad examples!

\end{displayquote}

In addition to these instructions, we provided the raters with demonstrations of the task. We include a few below as they were presented to the annotators:

\begin{displayquote}

\textbf{Example 1:}
\begin{itemize}

    \item  Category is “Socks”
\item Snippet is “This combination makes the fabric breathable, and durable with a comfortable grip”

\item Annotations: \begin{displayquote}
No implication
\end{displayquote} 
\end{itemize}

\textbf{Example 2:}
\begin{itemize}
    \item  Category is “Shirts \& Tops”
\item Snippet is “2019 New Trend Fashion Summer Solid Color Shirt Men Stand Collar Cotton Linen Half Sleeve Shirt Slim Type Male Casual Linen Tops”

\item Annotations: \begin{displayquote}
Bad example
\end{displayquote}  
\end{itemize}
\textbf{Example 3:}
\begin{itemize}
    \item  Category is “Quilts \& Duvets”
\item Snippet is “Egyptian cotton is soft and strong at the same time, making it durable.”

\item Annotations: \begin{displayquote}
Implies(Egyptian cotton, durable)\\
 Implies(Egyptian cotton, soft)\\
 Implies(Egyptian cotton, strong)\\
 Implies(soft and strong, durable)
\end{displayquote} 


\end{itemize}
\textbf{Example 4:}
\begin{itemize}
\item Category is “Tea \& Infusions”
\item Snippet is “Matcha green tea is more potent than regular green tea”

\item Annotations: \begin{displayquote}
Implies(Matcha green tea, more potent than regular green tea)
\end{displayquote} 
\end{itemize}

\textbf{Example 5:}
\begin{itemize}
    \item Category is “Handbags \& Purses”
\item Snippet is “It is made of polyester and nylon, making it durable and lightweight”

\item Annotations: \begin{displayquote}
Implies(polyester, durable) \\
Implies(nylon, durable) \\
Implies(polyester, lightweight) \\
Implies(nylon, lightweight)
\end{displayquote} 

Note: Even though the sentence says “and” here, we know that in the context of clothes, everything in the fabric has to have these properties. If unsure, the next best thing is to just say:
\begin{displayquote}

Implies(polyester and nylon, durable)\\
Implies(polyester and nylon, lightweight)

\end{displayquote}
\end{itemize}
\end{displayquote}

\section{Attribute splitting and typing annotator instructions}
\label{app:attribute-splitting-typing-instruction}

We provided raters with the following instructions for splitting attributes:

\begin{displayquote}

We use the nomenclature of attributes and types. A attribute is a specific value that describes a product, for example “red” or “affordable”. A type is a key (i.e. a name) that describes a collection of related attributes. For example, the attribute “red” belongs to the type of “color,” and the attribute “affordable” belongs to the type of “price.”
\begin{itemize}

\item The goal of this task is to get an exhaustive list of all the atomic attributes (attributes describing one simple aspect) relevant to the category contained in the snippet, along with type information for each attribute, and a short “canonical” summary of the attribute in the context of that type. “Canonical” here means that the summary you write in different snippets should be the same when describing a attribute with roughly the same meaning.

\item In this task, we will ask you to identify all of the attributes in a snippet of text relevant to a specific provided product category. If these attributes are associated with one of the provided previously annotated attributes, this should be specified. 

\item It might often be the case that a previously annotated attribute is not atomic, i.e. it describes a combination of more than one simple aspect of a product. In that case we want to annotate all those atomic attributes while also indicating that they correspond to a single compound previous attribute. 

\item In other cases, previously annotated attributes might include unnecessary extra text that becomes clear once types are annotated. In this case we also want to indicate that the new crisp, typed attribute corresponds to the previously annotated attribute.

\end{itemize}

\end{displayquote}

We provided raters with the following instructions for typing attributes:

\begin{displayquote}
A type is a named collection of attributes that describe a product. 
Examples: 
\begin{itemize}
    \item Attribute: Red > Type: Color
    \item Attribute: Glass > Type: Material
\end{itemize}
Naming a type might require world knowledge or additional searching the web. Some methods for type determination are listed below. If there are multiple applicable types, say a fine grained and a coarse grained one, list both comma separated.
\begin{itemize}
\item Read the snippet. Sometimes, the type can be found directly in the text. For example, to determine the type for the attribute “ivory” the snippet might just contain “ivory is a very traditional wedding dress color”.
\item Research the attribute on Google.com. For example, searching for [what is an ivory wedding dress] gives an answer that specifies that it is a color, not a material.
\item Some attributes might be very product-specific like “prevents skin from feeling sticky” in “this pillow prevents skin from feeling sticky.” In those cases, it’s ok to use a more vague type like “Benefits”. Vague types like “benefits” should be avoided when possible though. For example if a watch is listed as “waterproof”, we want a type like “Water-Resistance” and not simply “Benefits”, since “Water-Resistance” is something that people are often concerned about with watches.
\end{itemize}

\end{displayquote}

We also provided raters with several demonstrations of the task, an example of which is provided in Table \ref{tab:example-split}. Because these annotations are so structured, we provided them in a spreadsheet form.

\begin{table*}

    \centering
    \tiny
     \aboverulesep=0ex
 \belowrulesep=0ex
 \renewcommand{\arraystretch}{1.5}
\begin{tabular}{p{0.8cm}p{2cm}|lp{1.8cm}|llp{1.5cm}l} \hline
\toprule
\textbf{Category}                & \textbf{Snippet}                                                                                  & \textbf{Previous Tag}         & \textbf{Previous Summary}     & \textbf{New Facet}   & \textbf{New Tag}          & \textbf{New Summary} & \textbf{Associated Prev. Tag} \\ \midrule
 \multirow{4}{0.8cm}{Food Mixers \& Blenders} & 
 \multirow{4}{2cm}{Overall, the performance of Orient mixer grinder is good and best to use for small homes}  & Orient mixer grinder & Orient mixer grinder & Performance & good performance & good                  &                      \\ \cmidrule{3-8} 
                        &                                                                                          & for small homes      & small homes          & Type        & mixer grinder    & mixer grinder         & Orient mixer grinder \\ \cmidrule{3-8} 
                        &                                                                                          &                      &                      & Brand       & Orient           & Orient                & Orient mixer grinder \\ \cmidrule{3-8} 
                        &                                                                                          &                      &                      & Use Case    & for small homes  & small homes           & for small homes  \\
                        \bottomrule
\end{tabular}
\caption{Example annotation for splitting and typing task, as raters would see it.}
\label{tab:example-split}
\end{table*}

\end{document}